\documentclass{article}

\usepackage[final]{IIBPSL_2020}

\usepackage[utf8]{inputenc} 
\usepackage[T1]{fontenc}    
\usepackage{hyperref}       
\usepackage{url}            
\usepackage{booktabs}       
\usepackage{amsfonts}       
\usepackage{nicefrac}       
\usepackage{microtype}      
\usepackage{algorithm}
\usepackage{algorithmic}
\usepackage{amsmath}
\usepackage{graphicx}
\usepackage{url}            
\usepackage{xcolor}
\usepackage{appendix}
\usepackage{float}
\usepackage{caption}
\usepackage{sidecap}
\usepackage{natbib}
\usepackage{subfiles} 
\usepackage{pdfpages}

\usepackage{blindtext}
\usepackage{color}
\pagecolor{white}

\title{Graph Networks with Spectral Message Passing}

\author{
  Kimberly L. Stachenfeld \\
  DeepMind \\
  London, N1C 4AG \\
  \texttt{stachenfeld@google.com} \\
  \And
  Jonathan Godwin \\
  DeepMind \\
  London, N1C 4AG \\
  \texttt{jonathangodwin@google.com} \\
  \And
  Peter Battaglia \\
  DeepMind\\
  London, N1C 4AG \\
  \texttt{peterbattaglia@google.com} \\  
}
  
\begin{document}

\maketitle
\begin{abstract}
Graph Neural Networks (GNNs) are the subject of intense focus by the machine learning community for problems involving relational reasoning. 
GNNs can be broadly divided into spatial and spectral approaches. Spatial approaches use a form of learned message-passing, in which interactions among vertices are computed locally, and information propagates over longer distances on the graph with greater numbers of message-passing steps. Spectral approaches use eigendecompositions of the graph Laplacian to produce a generalization of spatial convolutions to graph structured data which access information over short and long time scales simultaneously. Here we introduce the Spectral Graph Network, which applies message passing to both the spatial and spectral domains. Our model projects vertices of the spatial graph onto the Laplacian eigenvectors, which are each represented as vertices in a fully connected ``spectral graph'', and then applies learned message passing to them. 
We apply this model to various benchmark tasks including a graph-based variant of MNIST classification, molecular property prediction on MoleculeNet and QM9, and shortest path problems on random graphs. Our results show that the Spectral GN promotes efficient training, reaching high performance with fewer training iterations despite having more parameters. The model also provides robustness to edge dropout and outperforms baselines for the classification tasks. We also explore how these performance benefits depend on properties of the dataset.
\end{abstract}

\section{Introduction}
\label{sec:intro}

Many machine learning problems involve data that can be represented as a graph, whose \textit{vertices} and \textit{edges} correspond to sets of entities and their relations, respectively.
These problems have motivated the development of graph neural networks (GNNs)~\citep{scarselli2008graph}, which adapt the notion of convolution on Euclidean signals to the graph domain~\citep{Bronstein_2017}. 
Here we introduce a GNN architecture which bridges two dominant approaches within the field---the spatial and spectral approach---to favorably trade-off their comparative strengths and weaknesses. 

Spatial approaches involve a form of learned message-passing~\citep{gilmer2017neural} that propagates information over the graph by a local diffusion process. 
Spectral approaches~\citep{bruna2013spectral} generalize the Fourier transform of Euclidean signals to graphs, providing access to information over short and long spatiotemporal scales simultaneously.
Spatial approaches have tended to be more popular recently; however, a limitation is that propagating information over long ranges can require many rounds of message-passing, resulting in fine-grained information being corrupted or lost.

To overcome this limitation, our Spectral Graph Network (GN) architecture performs message-passing over the input graph's structure---the ``spatial graph''---as well as message-passing in a high-level ``spectral graph''. 
This allows long-range information to be pooled, processed, and transmitted between any vertices in the spatial graph, which confers an inductive bias toward explicitly incorporating the global topology of the graph into its processing.
We test our Spectral GN on a graph MNIST classification task and on two distinct molecular property prediction tasks. Our model achieves high performances, more efficient training, and is more robust to dropped input vertices and edge sparsification in the model. These results demonstrate how spatial GNN approaches can benefit from low frequency information provided by spectral approaches.

\section{Related work}
\label{sec:related-work}

The field of GNNs has expanded rapidly in recent years, and a number of comprehensive reviews can be recommended~\citep{Bronstein_2017,gilmer2017neural,battaglia2018relational,zhou2018graph,wu2020comprehensive,zhang2020deep}. 

Spatial approaches to GNNs use learned, local filters defined over the local neighborhood of a vertex. These filters are applied across all vertices just as convolutional filters can be shared across a tensor.
The GraphNet (GN)~\citep{sanchez2018graph,battaglia2018relational} is a general formulation of the spatial approach to GNNs which can be parameterized to include message-passing neural networks (MPNNs)~\citep{gilmer2017neural}, graph convolutional networks (GCNs)~\citep{kipf2016semi}, and various others. We adopt the GN framework here as the representative of the spatial approach. GNs include a ``global'' term that uniformly pools features across the graph \citep{gilmer2017neural,battaglia2018relational}, though this is a crude mechanism because it pools into a single feature vector, without sub-structure. They are thus a good baseline against which to compare our more elaborate mechanism for handling low frequency information~\citep{gilmer2017neural,battaglia2018relational}. 
In contrast to hierarchical approaches which use a fixed hierarchy decided by the modeler~\citep{mrowca2018flexible,li2018learning}, our approach requires fewer domain-specific choices (e.g., how vertices comprise larger entities, the vertices' nearest neighborhood sizes, etc). Unlike the DiffPool model's~\citep{ying2018hierarchical} soft attention coarsening mechanism, our approach maintains the structure of the input graph throughout the computation.

Spectral approaches, which generalize the notion of spectral transforms of Euclidean signals to graphs~\citep{Bronstein_2017, ortega2018graph}, involve projecting the input signal onto $K$ eigenvectors of the graph Laplacian ordered by eigenvalue (Section~\ref{sec:graph-theory-background})---also known as taking the Graph Fourier Transform (GFT)---and applying filters over this fixed size representation. 
Spectral graph convolution has underpinned the rise of ``geometric deep learning'' \citep{Bronstein_2017}, where early work applied neural networks to the spectral representation of the input graph~\citep{bruna2013spectral}, and has led to graph CNNs via fast localized spectral filtering~\citep{defferrard2016convolutional}, CayleyNets~\citep{levie2018cayleynets}, and Graph Wavelet Neural Networks~\citep{xu2019graph}.
Spectral methods can offer advantages for computing global features via vertex pooling~\citep{ma2019graph,bianchi2019spectral}, learning globally sensitive vertex embeddings~\citep{you2019position}, and, most recently, for supporting directional message passing using the intrinsic low-dimensional geometry of the graph \citep{beaini2020directional}.
Unsupervised computation of the first $K$ eigenfunctions of the related Laplace-Beltrami operator can also be learned with deep learning~\citep{pfau2018spectral}.
A key challenge for spectral methods in deep learning is how to order the Laplacian eigenvectors. Ordering by eigenvalue can lead to degeneracies (when there are multiple equal eigenvalues), or instabilities (when slight perturbations of the graph cause changes in eigenvalue order). Our approach addresses this by representing eigenvectors through unordered vertices in the spectral graph.

\section{Model}

\subsection{Graph Theory Background}
\label{sec:graph-theory-background}
Let $G = (V, E)$ be a graph containing vertices $V$ and directed edges $E$. Let $v_i \in V$ be the vertex features for vertex $i$, $(e_k, r_k, s_k) \in E$ contain the edge features, sender indices, and receiver indices, respectively, and $g$ be graph-level ``global'' features.
The adjacency matrix, $A$, is defined such that $A_{ij}=1$ if $(\cdot, i, j) \in E$ and $0$ otherwise. The degree matrix $D$ is diagonal with $D_{ii} = \sum_j A_{ij}$.

The Laplacian is a positive semi-definite matrix $L=D-A$ that describes diffusion over a graph  \citep{Chung1994}.
\footnote{Note, this is defined for simple graphs. Our graphs are not simple, since they can be directed and contain loops; however, all of the graphs in our dataset are symmetric and do not contain self-loops, so we can treat them as simple for this step. See~\citep{Singh_2016} for generalizations of spectral methods to directed graphs.}
The eigendecomposition of $L$ is $U \cdot \mathrm{diag}(\Lambda) \cdot U^\top$, where $U$ is the matrix of $|V|$ eigenvectors and $\Lambda$ vector of eigenvalues. The operation $U^\top \phi$ projects a signal $\phi$ over graph vertices into the spectral domain.
The $K$ eigenvectors with the smallest eigenvalues distinguish vertices that will be slowest to share information under diffusion (or, similarly, message passing) \citep{shi2000normalized}.
Other applications include spectral clustering \citep{ng2002spectral} and graph filtering \citep{ortega2018graph}.
Intuitively, these eigenvectors will therefore allow us to bridge vertices that message passing will most struggle to integrate given the graph structure.
Projecting vertex latents onto these eigenvectors and processing the projected signal can also be thought of as augmenting message passing with a nonlinear, message-passing based low-pass filter.

\subsection{Our Spectral GN model}
\label{sec:sgn-model}

\begin{figure}[t]
  \centering
  \includegraphics[width=\columnwidth]{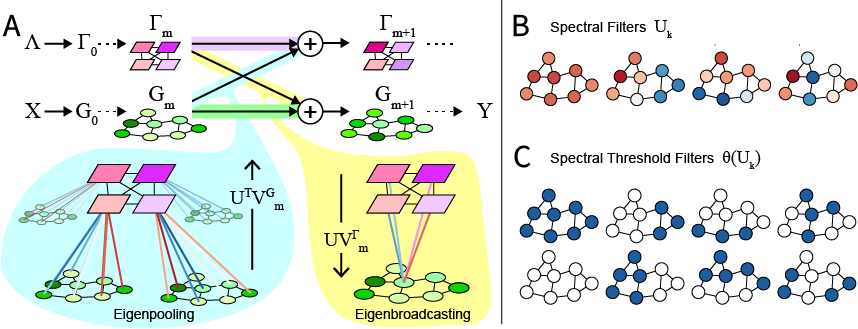}
  \caption{\label{fig:schematic} \textbf{(A)} Spectral GraphNet Schematic.
  The spatial graph, $G_m$, is processed by the GNN$^G$ network (green~graphs~and~arrow). The spectral graph, $\Gamma_m$, is processed in parallel by the GNN$^\Gamma$ network (purple~graphs~and~arrow). 
  The eigenpooling operation (cyan~bubble) communicates $G_m$'s vertex information to $\Gamma_{m+1}$, weighted by the eigenvector values (blue/red lines). The eigenbroadcasting operation (yellow~bubble) communicates $\Gamma_m$'s vertex information to $G_{m+1}$, weighted by the eigenvector values. \textbf{(B)} Spectral Filters $U_{K=4}$ (first 4 non-thresholded eigenvectors). \textbf{(C)} Thresholded Spectral Filters $\theta(U_{K=4})$, first 4 eigenvectors $U_{K=4}$ (top row), and their negatives, $-U_{K=4}$ (bottom row), thresholded at 0. Vertex coloring indicates weights on each spatial latent applied before pooling.
  }
\end{figure}

\textbf{Spatial and Spectral GraphNets.} 
The input ``spectral graph'' is a complete graph with $K$ vertices corresponding to the $K$ smallest eigenvalues of input spatial graph $X$'s graph Laplacian. The vertex features are initialized to the eigenvalues, $\Lambda_{:K}$. The spatial and spectral input graphs are processed by vertex and edge-wise MLP encoders to yield $G_0$ and $\Gamma_0$, respectively.

On the $m$-th message passing step, spatial and spectral GNNs, GNN$^G$ and GNN$^\Gamma$, are applied to $G_m$ and $\Gamma_m$, respectively (horizontal lines between $m$ and $m+1$ steps in Figure~\ref{fig:schematic}A).
After $M$ rounds of message passing, an MLP decoder processes $G_M$ and returns graph output $Y$. For graph-level classification, the loss is applied to the global feature of $Y$. 

For the GNNs, we implemented the GN, Graph Convolution Network (GCN), and (for spectral processing only) the Graph Fourier Transform (GFT).
The GCN is a popular, lightweight spatial approach that has fewer parameters and no global term. 
The GFT, inspired by~\citep{bruna2013spectral}, applies an MLP to the spectral latents ordered by eigenvalue.

\textbf{Eigenpooling/broadcasting.} 
vertex features of $G_m$ are projected onto the $k$-th vertex in the spectral domain via multiplication with the $k$-th eigenvector $U_{\cdot, k}$, or, in matrix form, $\bar{V}^\Gamma_{m+1} = U^\top V^G_m$ (Figure~\ref{fig:schematic}A, blue). For notational simplicity, we let $U$ be the truncated $K$ eigenvector matrix.
These eigenpooled vertices, $\bar{V}^\Gamma_{m+1}$, are concatenated onto the spectral vertices, $\hat{V}^\Gamma_{m+1}$, to form $V^\Gamma_{m+1}$.
Similarly, vertex features of $\Gamma_m$ are projected to the $i$-th vertex in the spectral domain via multiplication with $U_{i, \cdot}$, or, in matrix form, $\bar{V}^G_{m+1} = U V^\Gamma_m$ (Figure~\ref{fig:schematic}A, yellow). 
These eigenbroadcasted spectral vertices, $\bar{V}^G_{m+1}$, are concatenated onto the spatial vertices, $\hat{V}^G_{m+1}$, to form $V^G_{m+1}$.

We also explored a modified variant termed ``Spectral Threshold GN'' ($\theta(U)$-GN), which thresholds $U$ at 0, and uses $\texttt{concat}[\theta(U), \theta(-U)]$ as the projection matrix (Figure~\ref{fig:schematic}B,C). It thus has $2 K$ the number of spectral graph vertices, whose input for initialization are the duplicated eigenvalues. The $U$-GN refers to a Spectral GN with no thresholding.

\textbf{Further considerations.} 
By treating the eigenvectors as an unordered set \emph{labelled} by eigenvalue, our approach helps circumvent instabilities and degeneracies that challenge previous approaches which treat them as a sequence \emph{ordered} by eigenvalue. 
If we use $K=1$, this is analogous to using a global term for graph-level communication~\citep{gilmer2017neural,battaglia2018relational}. 
%

Compared to other hierarchical GNN schemes~\citep{mrowca2018flexible,li2018learning,ying2018hierarchical}, Spectral GN uses the matrix of eigenvectors to exchange low- and high-level vertex information.
Intuitively, the spectral augmentation can also be thought of as nonlinear low pass filtering on learned latents \citep{ortega2018graph}.
Eigendecompositions can be expensive ($\mathcal{O}(N^3)$) to compute; however, approximations \citep{hammond2009wavelets} and learned models \citep{pfau2018spectral} can mitigate this expense.


\section{Experiments}

\begin{SCfigure}
  \centering
  \includegraphics[width=.5\textwidth]{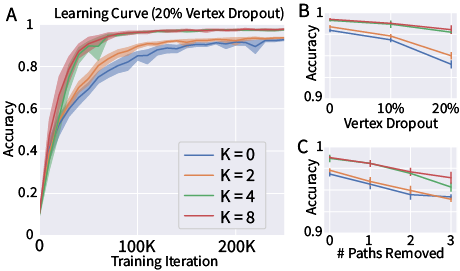}
  \caption{\label{fig:mnist-main-results} Graph MNIST results. \textbf{(A)} Learning curves showing classification test accuracy ($y$-axis) across training iterations ($x$-axis). \textbf{(B)} Accuracy ($y$-axis) of $U$-GN with increasing vertex dropout proportion ($x$-axis). \textbf{(C)} Accuracy ($y$-axis) of $U$-GN with increasing number of shortest paths removed ($x$-axis).} 
\end{SCfigure}

We evaluated our models and baselines on three graph property prediction task, Graph-MNIST~\citep{defferrard2016convolutional}, MoleculeNet-HIV molecule classification~\citep{Wu_2018}, and QM9 quantum molecular property prediction~\citep{ramakrishnan2014quantum}, as well as one vertex property prediction tas, shortest path computation.
Across these benchmarks, we found that our hybrid spectral architectures yielded efficient training, were more robust to both missing vertices in the inputs and pruned edges during message passing, and in three cases (MNIST, MoleculeNet-HIV, shortest path on 2D random  graphs) yielded higher overall performance. 
For molecular tasks, the thresholded Spectral GNs outputperformed nonthresholded, while for MNIST and shortest path problems, the non-thresholded ones were.
Additional information, including tables of numerical results, learning curves, training specifics, and dataset details can be found for all experiments in the Appendix.

\subsection{Graph MNIST}
\label{sec:mnist}

MNIST handwritten digit classification~\citep{lecun1998gradient} can be adapted for graphs by treating each pixel as a vertex and joining neighboring pixels with an edge \citep{defferrard2016convolutional}. 
Each sample consists of a $28\times28$ grid and and edges join the four axis-aligned neighbors.

Since we wanted to assess the ability of spectral message passing to process information across large, sparse graphs, we did not use superpixels \citep{defferrard2016convolutional, dwivedi2020benchmarking}.
Vertex features were the pixels' intensities and edge features contained the 2D displacement vector from sender to receiver vertex position.
Under ``uniform vertex dropout'', vertices were removed uniformly at random from each graph ($p_\text{dropout}\in[0, 0.1, 0.2]$). 
Under ``shortest path vertex dropout'', pairs of vertices were randomly selected, and all of the vertices along one of the shortest paths connecting them were removed ($n_\text{paths}\in[0, 1, 2, 3])$.
This creates contiguous holes that substantially increase the graph's diameter, (i.e., the length of the longest shortest-path distance between any two vertices), 
challenging approaches that rely strictly on local message passing and altering the first $K$ eigenvectors.

Our $U$-GN reached the highest performance, trained more efficiently, and was more robust to vertex dropout (Figure \ref{fig:mnist-main-results}). The $U$-GN spectral approach ($0.992$) outperformed GCN ($0.8451$), GN-GFT ($0.925$), and GCN-GFT ($0.925$) for 0 dropout, and outpeformed thresholded $\theta(U)$-GN ($0.978$). The GCNs deteriorated rapidly under dropout, while GN-GFTs deteriorated specifically under shortest path dropout.
However, the vanilla GN showed a greatest decrement in performance compared with the Spectral GNs under vertex dropout (Figure~\ref{fig:mnist-main-results}B). 
$U$-GN ($K=4$ or $K=8$) was the highest performing model under random shortest path vertex dropout (Figure~\ref{fig:mnist-main-results}C). However, increased dropout proportions appeared to weaken the $U$-GN and vanilla GN comparably, suggesting these perturbations to the global structure were harder for the Spectral GNs to overcome.

\subsection{Molecular Property Prediction (MoleculeNet-HIV and QM9)}
\label{sec:molecular}

\begin{figure}
  \includegraphics[width=\columnwidth]{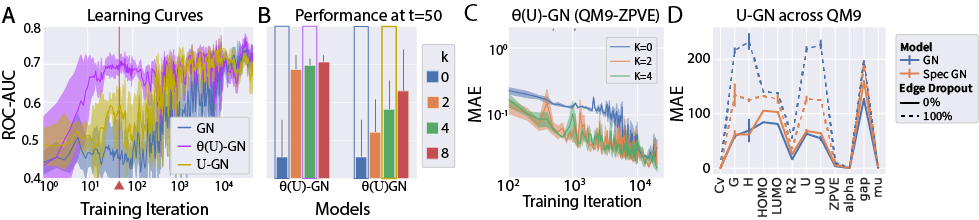}
  \caption{\label{fig:molecular} Molecular classification results. 
  \textbf{(A)} Learning curves showing ROC-AUC ($y$-axis) across training iterations ($x$-axis) for GN, $U$-GN, and $\theta(U)$-GN. Red line indicates $t=50$ training iterations. 
  \textbf{(B)} ROC-AUC across all spectral models, with $U$ and $\theta(U)$ augmentation, sampled at $t=50$ iterations. Models shown in (A) marked with boxes whose outline colors match their colors in (A).
  \textbf{(C)} Learning curves showing test loss early in training for an example QM9 target, ZPVE. Blue lines are networks without spectral information, Orange is $\theta(U)$ $(k=2)$ and Green is $\theta(U)$ $(k=4)$. Spectral models show better performance early on in training.
  \textbf{(D)} Final test MAE in target units, with best performing spectral model displayed per edge dropout value. Target units in Appendix.
  } 
\end{figure}

MoleculeNet-HIV involves predicting whether a molecule can inhibit HIV replication from its molecular graph~\citep{Wu_2018,hu2020open}. 
QM9 is a quantum chemistry benchmark that that involves predicting 13 target properties from the annotated molecular graph \citep{ramakrishnan2014quantum}. 
As is standard, we use separately trained models for each target and target whitening \citep{gilmer2017neural}. 

For both benchmarks, thresholded $\theta(U)$ augmented spectral models trained more efficiently than other models.
The $\theta(U)$-GNs $(k=4)$ reached 87\% of state of the art in \textasciitilde60 iterations (1920 samples), while GCN and GN took \textasciitilde500 and \textasciitilde10,000 iterations, respectively. For QM9, we found that the spectral models' MAE was lower for the first \textasciitilde20K steps of training (Figure~\ref{fig:molecular}C). 

Final performance on MoleculeNet-HIV was comparable to the top GNN methods on the Open Graph Benchmark leaderboard, GCN+GraphNorm (ROC-AUC=$0.7883$)~\citep{hu2020open,dwivedi2020benchmarking}. Our GNs reached scores of $0.739\pm0.029$, $U$-GN $0.753\pm0.028$, and $\theta(U)$-GN $(K=4)$ $0.769\pm0.015$.
For QM9, the vanilla GN was the top performing model among those were evaluated. Spectral GCNs did outperform vanilla GCNs. 
Our models were neither molecule-specific nor tailored to QM9, and did not reach state of the art performance, generally having 2--10$\times$ SOTA MAE.

GNNs typically train more efficiently but perform worse with sparsified edges. We tested whether message passing over the spectral graph could potentially compensate for edge dropout of 50\% and 100\% to allow more efficient procesesing.
%
We found that spectral significantly outperformed their vanilla counterparts under 100\% edge removal (Figure~\ref{fig:molecular}D, orange-dashed $U$-GN outperforming blue-dashed GN for 12/13 targets), but did not reach the solid lines denoting the non-sparsified graph. For 50\% edge dropout: $U$-GN had lowest error for 5/13 targets, and for 0\%, 2/13.

\subsection{Shortest Path Prediction on Random Graphs}
\label{sec:shortestpath}

\begin{figure}
  \includegraphics[width=\columnwidth]{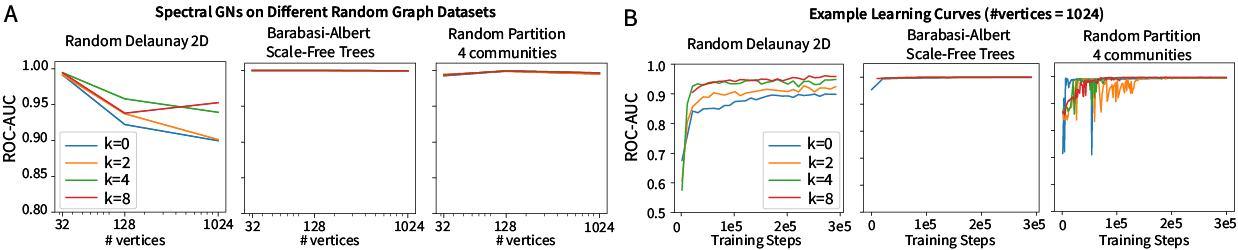}
  \caption{\label{fig:shortestpath} Shortest path prediction on random graphs. 
  \textbf{(A)} Performance of $U$-GNs and GN ($k=0$) on for different random graph datasets and varying numbers of vertices. 
  \textbf{(B)} Learning curves of $U$-GNs and GN ($k=0$) for 1024 vertices.
  } 
\end{figure}

For each of the random graph datasets, 5000 random graphs and pairs of vertices were sampled, and the network was tasked with predicting vertices that lie on the shortest path joining the vertex pair. The random graph datasets included Barab\'{a}si-Albert scale-free tree graphs, Random Partition graphs, Random Delaunay 2D graphs, which  were sampled by distributing points uniformly at random in a square and connected according to a Delaunay triangulation, and Random Delaunay 2D graphs with shortest path dropout as described in Section\ref{sec:mnist} to introduce obstacles. These capture different properties of graphs observed in real-world graph data: scale-free (Barab\'{a}si-Albert), community structure (Random Partition), and low dimensionality (Random Delaunay 2D).
See Appendix for more details.

On 2D Random Delaunay graphs, $U$-GNs ($K>0$) increasingly outperform GN ($K=0$) on shortest path prediction as number of vertices increase (Figure~\ref{fig:shortestpath}A). This trend appears in the path dropout conditions as well (see Appendix). 
This is consistent with performance benefits on MNIST graphs, which have similar statistics. 
This effect is not seen, however, on Barab\'{a}si-Albert or random partition graphs, on which all models perform similarly (Figure~\ref{fig:shortestpath}A). 
For the Barab\'{a}si-Albert graphs, all models reach high performance quickly compared to the other datasets (Figure~\ref{fig:shortestpath}B). For random partition graphs, we see that spectral augmentation appears to make learning \emph{less} efficient, although ultimately all models perform similarly. This may be partly because cluster position is inferrable from local neighborhoods, making global information redundant; because low-frequency eigenvectors are promoting over-smoothing for this dataset; or because the higher frequency eigenvectors do not contain meaningful information given the graphs have 4 clusters. These results demonstrate that the ability of spectral augmentation to be useful for various other real world problems is likely to depend on the size and statistics of the graphs involved. 

We looked at the contribution of spatial and spectral message passing by replacing the spectral and spatial message passing components with components process the node features without using edge or global pooling to share information among nodes (see Appendix). 
We see that removing spatial message passing negatively affects performance in all cases datasets. For Random Delaunay 2D graphs, spatial and spectral message passing slightly but consistently the spectral message passing ablations. This suggests that much of the benefit of the spectral augmentation is from processing spectral node features, and that interactions among the eigenvectors play a smaller role.

\section{Conclusion}
We introduce Spectral GraphNets, which combine spatial and spectral GNNs. Across our experiments, we find that the Spectral GNs reach competitive performance, train efficiently, and compensate for missing vertices in the data and edge dropout in the GN.
Our results support the view low-frequency spectral components contain information that is relevant to certain problems on graphs that spatial GNNs have difficulty accessing for certain types of graph problems.

\bibliographystyle{plainnat}
\bibliography{references}




\section*{Supplementary material}

# Appendices

## A  Implementation

Models were trained with learning rate of $10^{-4}$ for 250K steps (Graph MNIST), $5 \cdot 10^{-4}$ for 50K steps (MoleculeNet-HIV), $5 \cdot 10^{-4}$ for $10^7$ steps (QM9), or $3 \cdot 10^{-5}$ for $3 \cdot 10^5$ steps. The encoders, decoders, node, edge, and global updating functions all consisted of MLPs with 3 layers, 32 hidden units, and ReLU activation functions. Sum pooling was used to pool messages, and layer norm was applied to the output of the encoder and all message passing networks, and the activation function was not applied to the last layer of the networks performing message passing. The update functions were further modified to access the history of latents: at each round of message passing, initial, previous, and current latents were concatenated. We also uses residual connections, meaning that the current and previous latent were summed after each step of message passing for each feature. We found this to provide a slight edge in training (though, anecdotally, did not see qualitatively different effects). Empty edge or node features in the input were initialized to 1.

The total number of parameters varies between GN, GCN, GN-GFT, and GCN-GFT because these MLPs are composed in different ways. GCNs do not have globals or edge updating functions. For the GFT (Graph Fourier Transform) spectral core, spectral nodes were ordered by eigenvalue index and reshaped into a $K \times n_\text{latents}$ vector, then fed processed by an MLP. Rather than repeating the same GN or GCN for each iteration of message passing, a unique core was used for each step. This permits each step of message passing to apply a different function.

## B  Dataset Overview

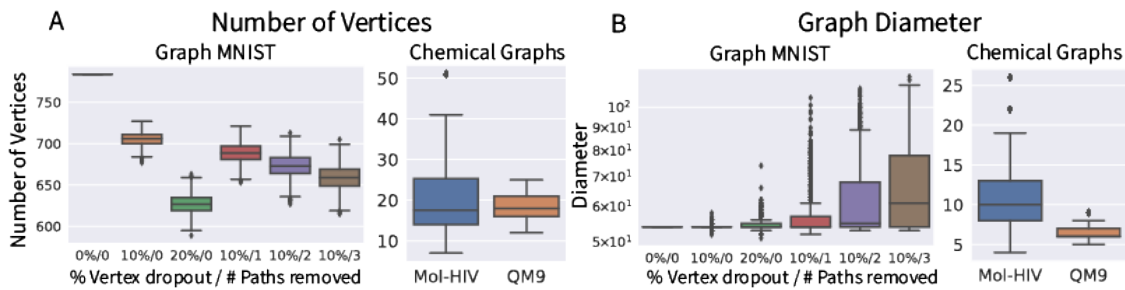

Figure B.1: **(A)** Distribution of number of nodes per graphs across tasks **(B)** Distribution of diameter per graphs across tasks

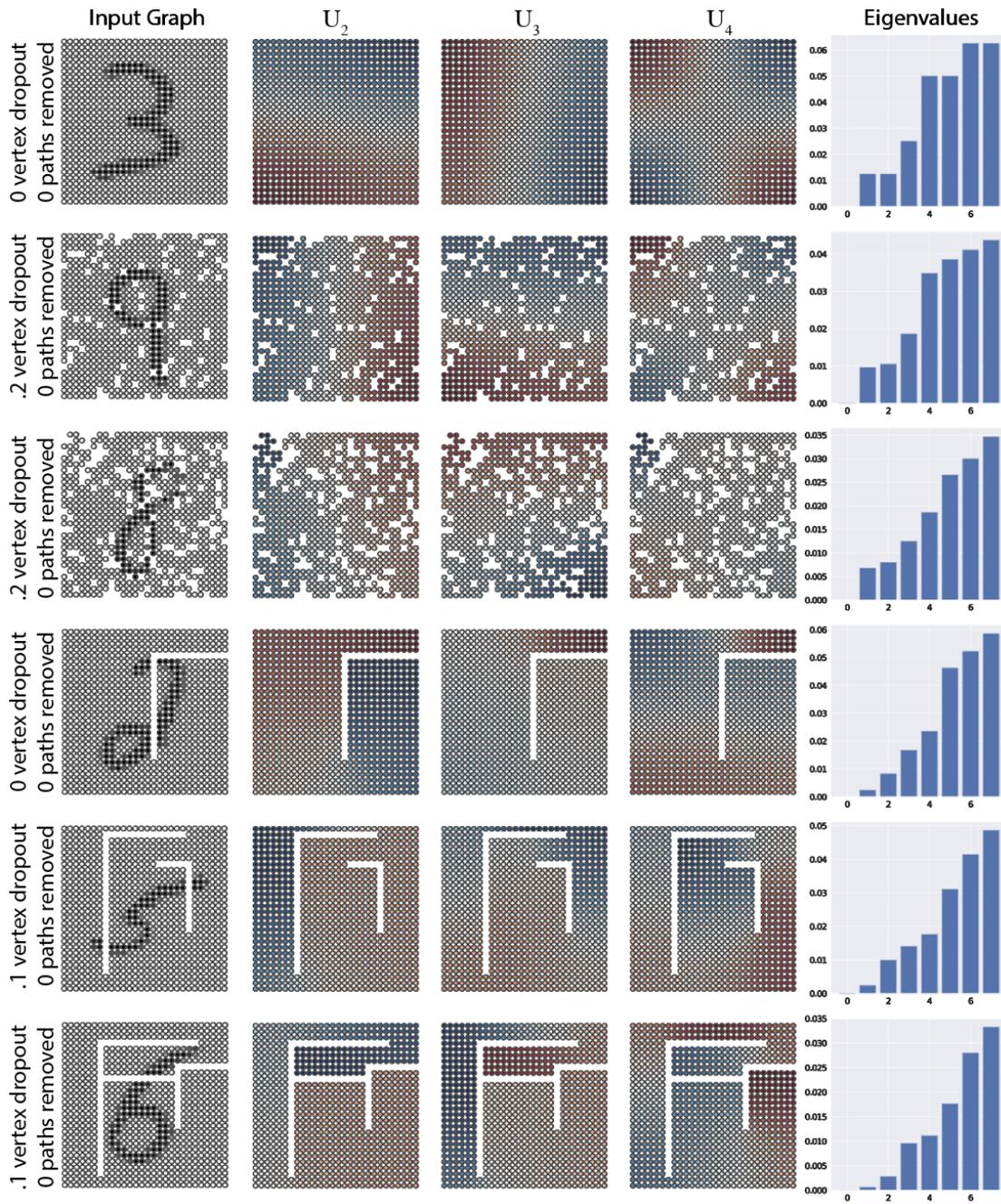

Figure B.2: Samples from Graph MNIST with various levels of dropout. First 4 Laplacian eigenvectors and eigenvalues are shown.

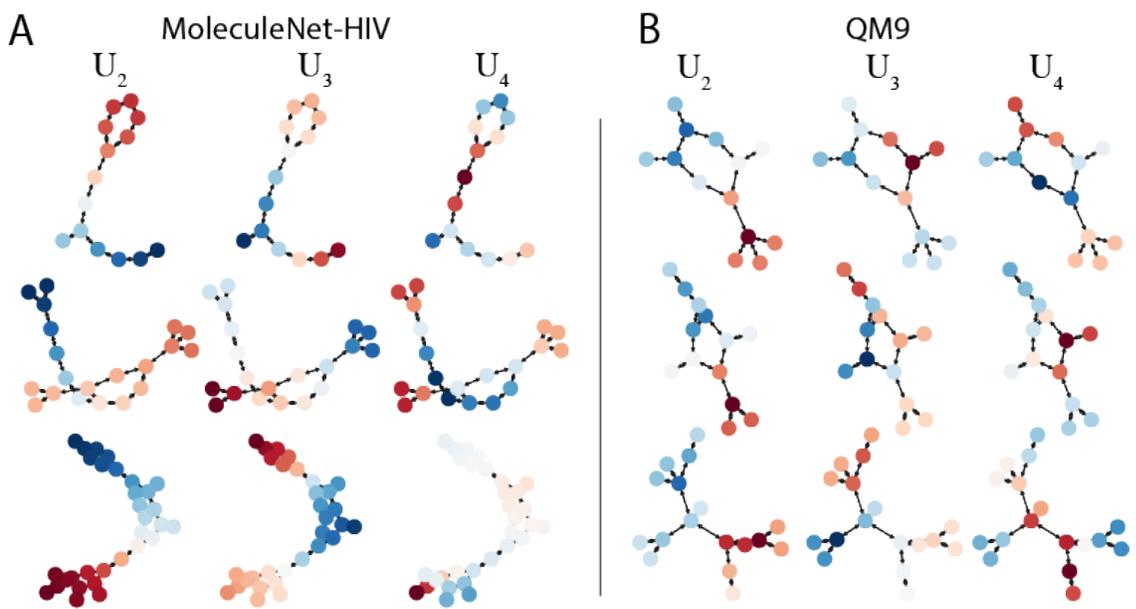

Figure B.3: Samples from (A) MoleculeNet-HIV and (B) QM9 with first 3 nontrivial Laplacian eigenvectors.

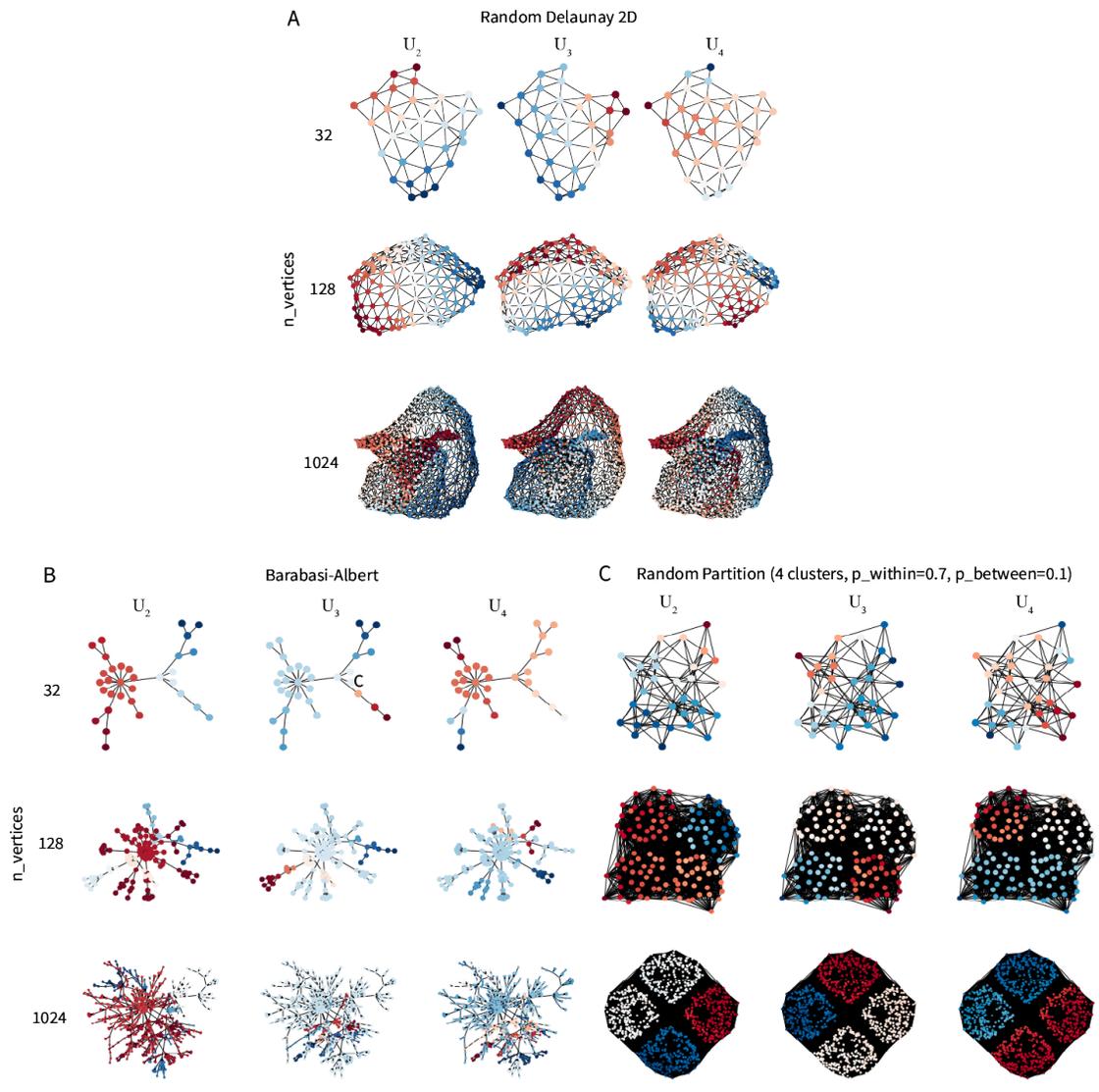

Figure B.4: Samples and eigenvectors from (B) Random Delaunay 2D Graphs, (B) Barabási-Albert ($m = 1$) random graphs, and (C) Random Partition Graphs.

# C MNIST

## C.1 MNIST Learning Curves

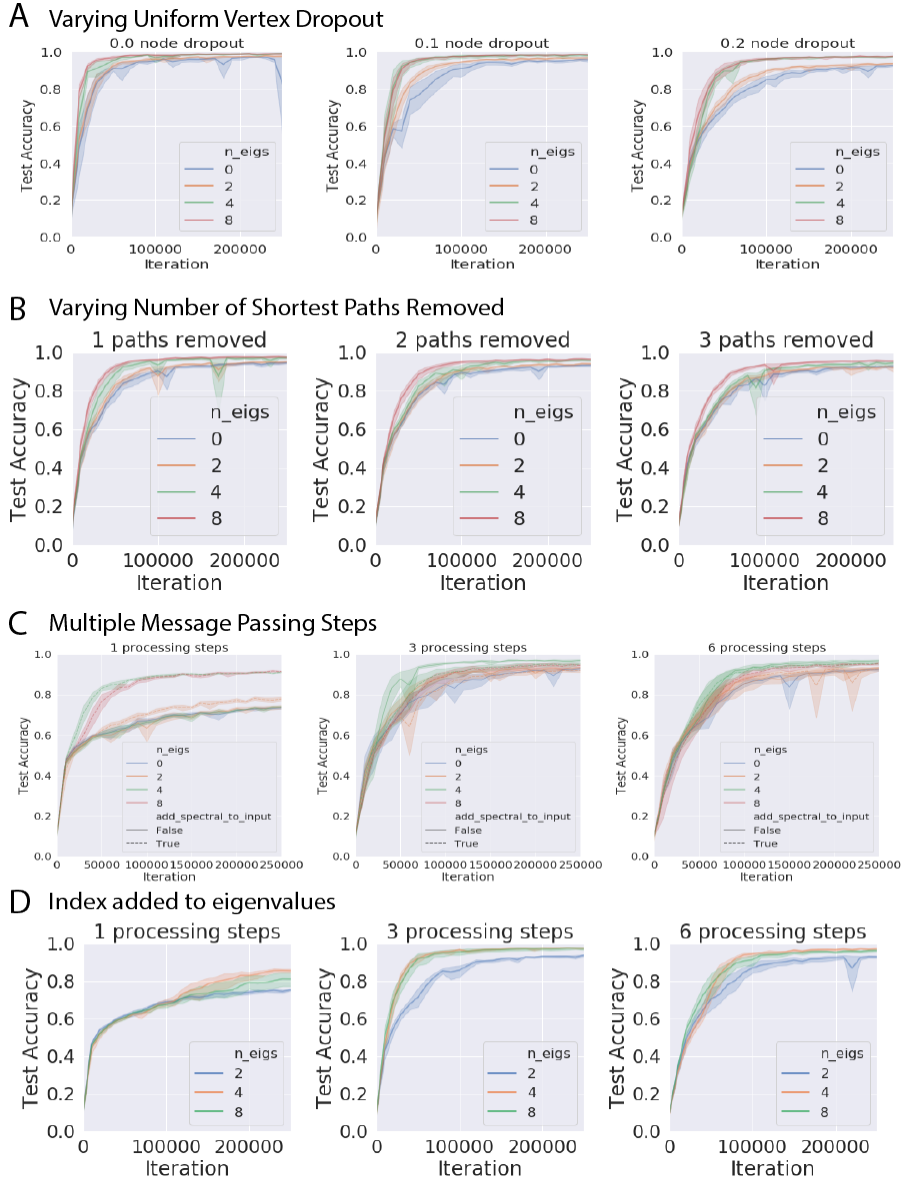

Figure C.5: Learning curves for GN and spectral $U$-GNs. Curves depict mean and standard deviation for each model type averaged over 5 seeds. Unless otherwise noted, runs have 0 shortest path dropout and 3 steps of message passing. **(A)** Varying rates of node dropout. **(B)** Varying numbers of shortest paths removed (node dropout rate = 0.1). **(C-D)** Varying numbers of message passing iterations (node dropout rate = 0.2). In (C), solid line indicates $U$-GN, dashed line GN+$U$ appended to node inputs. In **(D)**, all runs use $U$-GN with eigenvalue index appended to eigenvalues in the inputted spectral graph.

**MNIST: Message passing iterations**  Since the spectral models should, in principle, more easily aggregate information across large distances, we evaluated how the spectral models compared to simply increasing the number of message passing steps (using a uniform node dropout rate of 0.2).

As shown in Appendix Table C.3, and Figures C.5 and 2C,D, for 1 message passing step, the non-spectral GN, $U$-GN, and $\theta(U)$-GN performed poorly, with accuracy of only $0.748 - 0.753$ in 250K steps.

5

For 3 message passing steps, the spectral $U_4$-GN and $U_8$-GN had highest performance, with accuracy of 0.98 compared to the non-spectral GN's 0.94. We expected increasing the number of message passing steps to permit GN to approach the performance of $U$-GN. However, for 6 message passing steps, the $U$-GN still reached higher performance earlier in training than GN (see Appendix Figure C.5) and after 250K steps, the non-spectral GN was still behind with an accuracy of 0.938 compared to $U$-GN's 0.976 ($K = 4$). However, these models train considerably slower in terms of number of training iterations and wall clock time, so it is possible this may have changed with evern more training time.

There were two experimental conditions under which a spectral model was able to excel with only 1 message passing iteration. The first was to simply append $U$ to the inputted graph's node features. This suffers from degeneracies/instabilities since eigenvectors are ordered by index (Section 2) and does not permit the eigenvectors to explicitly route messages. However, this does permit nodes to access global information. Generally, integrating the eigenvectors via spectral message passing had better performance than providing the eigenvectors as input (Appendix Table C.3; Figures 2B-C). However, for only 1 message passing step, the appended eigenvectors significantly outperformed the spectral message passing model.

The other condition was to append node index information to the incoming spectral graph. Typically, the node latents of the spectral graph are set to the eigenvalue corresponding to each node. We also experimented with appending a one-hot indicating the index of that eigenvector. In theory, this should be easy for the spectral graph to deduce with a small number of message passing iterations. However, in the absence of a large number of message passing steps, appending eigenvalue index boosted learning for 1 but had no discernible effect for 3 or 6 message passing iterations (Appendix Table C.3).

## C.2 MNIST Tables

| | | 0% node Dropout | | | |
|---|---|---|---|---|---|
| | $K$ | 0 | 2 | 4 | 8 |
| $U$ | GCN | 0.846±0.010 | 0.849±0.007 | **0.851±0.004** | 0.851±0.005 |
| | GCN-GFT | 0.846±0.010 | 0.882±0.009 | **0.925±0.007** | 0.867±0.023 |
| | GN | 0.979±0.002 | 0.983±0.002 | 0.991±0.001 | **0.992±0.001** |
| | GN-GFT | 0.979±0.002 | 0.984±0.002 | **0.987±0.002** | 0.985±0.002 |
| $\theta(U)$ | GCN | **0.846±0.010** | 0.571±0.022 | 0.561±0.019 | 0.564±0.010 |
| | GCN-GFT | **0.846±0.010** | 0.551±0.013 | 0.599±0.028 | 0.580±0.005 |
| | GN | 0.979±0.002 | 0.977±0.003 | **0.978±0.003** | 0.947±0.045 |
| | GN-GFT | **0.979±0.002** | 0.935±0.010 | 0.951±0.011 | 0.923±0.029 |
| | | 10% node Dropout | | | |
| $U$ | GCN | 0.694±0.007 | **0.696±0.006** | 0.694±0.006 | 0.691±0.009 |
| | GCN-GFT | 0.694±0.007 | 0.730±0.007 | **0.782±0.028** | 0.638±0.040 |
| | GN | 0.968±0.003 | 0.973±0.002 | 0.987±0.004 | **0.987±0.002** |
| | GN-GFT | 0.968±0.003 | 0.973±0.004 | **0.982±0.004** | 0.977±0.003 |
| $\theta(U)$ | GCN | **0.694±0.007** | 0.467±0.006 | 0.467±0.009 | 0.466±0.005 |
| | GCN-GFT | **0.694±0.007** | 0.425±0.013 | 0.442±0.016 | 0.439±0.019 |
| | GN | 0.968±0.003 | 0.967±0.004 | 0.967±0.011 | **0.974±0.009** |
| | GN-GFT | **0.968±0.003** | 0.915±0.045 | 0.905±0.058 | 0.828±0.030 |
| | | 20% node Dropout | | | |
| $U$ | GCN | 0.566±0.008 | **0.567±0.007** | 0.561±0.010 | 0.566±0.011 |
| | GCN-GFT | 0.566±0.008 | 0.565±0.016 | **0.605±0.014** | 0.456±0.024 |
| | GN | 0.940±0.005 | 0.950±0.004 | 0.977±0.002 | **0.980±0.005** |
| | GN-GFT | 0.940±0.005 | 0.946±0.004 | **0.970±0.008** | 0.948±0.007 |
| $\theta(U)$ | GCN | **0.566±0.008** | 0.424±0.017 | 0.419±0.008 | 0.406±0.021 |
| | GCN-GFT | **0.566±0.008** | 0.358±0.012 | 0.351±0.023 | 0.372±0.009 |
| | GN | 0.940±0.005 | 0.940±0.004 | 0.941±0.020 | **0.945±0.015** |
| | GN-GFT | **0.940±0.005** | 0.874±0.060 | 0.817±0.060 | 0.738±0.063 |

Table C.1: (Above) Overall best test classification accuracy across 5 seeds given 2.5e5 training steps for different GraphNet architectures given different levels of uniform node dropout.

7

| | 0 paths removed, 10% node dropout | | | | |
|---|---|---|---|---|---|
| | $K$ | 0 | 2 | 4 | 8 |
| $U$ | GCN | 0.694±0.007 | **0.696±0.006** | 0.694±0.006 | 0.691±0.009 |
| | GCN-GFT | 0.694±0.007 | 0.730±0.007 | **0.782±0.028** | 0.638±0.040 |
| | GN | 0.968±0.003 | 0.973±0.002 | 0.987±0.004 | **0.987±0.002** |
| | GN-GFT | 0.968±0.003 | 0.973±0.004 | **0.982±0.004** | 0.977±0.003 |
| $\theta(U)$ | GCN | **0.694±0.007** | 0.467±0.006 | 0.467±0.009 | 0.466±0.005 |
| | GCN-GFT | **0.694±0.007** | 0.425±0.013 | 0.442±0.016 | 0.439±0.019 |
| | GN | 0.968±0.003 | 0.967±0.004 | 0.967±0.011 | **0.974±0.009** |
| | GN-GFT | **0.968±0.003** | 0.915±0.045 | 0.905±0.058 | 0.828±0.030 |
| | 1 path removed, 10% node dropout | | | | |
| $U$ | GCN | 0.625±0.019 | **0.646±0.001** | 0.641±0.003 | 0.631±0.015 |
| | GCN-GFT | 0.625±0.019 | **0.632±0.009** | 0.599±0.018 | 0.511±0.011 |
| | GN | 0.957±0.006 | 0.960±0.005 | **0.981±0.000** | 0.981±0.004 |
| | GN-GFT | 0.957±0.006 | 0.956±0.007 | **0.973±0.001** | 0.954±0.006 |
| $\theta(U)$ | GCN | **0.625±0.019** | 0.446±0.006 | 0.436±0.005 | 0.446±nan |
| | GCN-GFT | **0.625±0.019** | 0.362±0.034 | 0.397±0.032 | 0.400±0.018 |
| | GN | **0.957±0.006** | 0.914±0.052 | 0.941±nan | 0.955±0.012 |
| | GN-GFT | **0.957±0.006** | 0.844±0.085 | 0.862±0.088 | 0.762±0.032 |
| | 2 path removed, 10% node dropout | | | | |
| $U$ | GCN | 0.595±0.009 | **0.605±0.004** | 0.595±0.006 | 0.593±0.008 |
| | GCN-GFT | **0.595±0.009** | 0.583±0.002 | 0.525±0.005 | 0.470±0.014 |
| | GN | 0.945±0.006 | 0.949±0.005 | 0.969±0.001 | **0.971±0.005** |
| | GN-GFT | 0.945±0.006 | 0.952±0.001 | **0.957±0.007** | 0.934±0.010 |
| $\theta(U)$ | GCN | **0.595±0.009** | 0.438±0.015 | 0.430±nan | 0.423±0.004 |
| | GCN-GFT | **0.595±0.009** | 0.385±0.013 | 0.353±0.021 | 0.365±0.023 |
| | GN | **0.945±0.006** | 0.943±0.002 | 0.837±0.205 | 0.936±0.014 |
| | GN-GFT | **0.945±0.006** | 0.788±0.007 | 0.768±0.023 | 0.748±0.009 |
| | 3 path removed, 10% node dropout | | | | |
| | $K$ | 0 | 2 | 4 | 8 |
| $U$ | GCN | 0.567±0.007 | 0.579±nan | 0.559±0.014 | **0.568±0.009** |
| | GCN-GFT | 0.567±0.007 | 0.559±0.005 | 0.483±0.007 | 0.427±0.020 |
| | GN | 0.942±0.003 | 0.939±0.003 | 0.953±0.005 | **0.964±0.007** |
| | GN-GFT | 0.942±0.003 | 0.936±nan | **0.949±0.006** | 0.922±0.009 |
| $\theta(U)$ | GCN | **0.567±0.007** | 0.414±0.007 | 0.407±nan | 0.408±0.001 |
| | GCN-GFT | **0.567±0.007** | 0.366±0.008 | 0.340±0.010 | 0.343±0.015 |
| | GN | 0.942±0.003 | 0.936±nan | **0.951±nan** | 0.918±0.024 |
| | GN-GFT | 0.942±0.003 | **0.945±nan** | 0.816±0.104 | 0.822±0.099 |

Table C.2: (Above) Overall best test classification accuracy across 5 seeds given 2.5e5 training steps for different GraphNet architectures. Varying levels numbers of shortest paths removed.

|   |    |              | 1 processing steps, 20% node Dropout |                   |                   |                   |
|---|----|--------------|--------------------|-------------------|-------------------|-------------------|
|   |    | $K$          | 0                  | 2                 | 4                 | 8                 |
| $U$ | GN | spectral MP | 0.746±0.003       | **0.749±0.007**  | 0.748±0.004      | nan±nan           |
|   |    | as input     | 0.746±0.003       | 0.795±0.005      | **0.915±0.004** | 0.921±0.006      |
|   | GN | index eigvals | 0                 | 0.764±nan        | 0.877±nan        | **0.864±0.023** |
|   |    |              | 3 processing steps, 20% node Dropout |               |                  |                   |
| $U$ | GN | spectral MP | 0.940±0.005       | 0.950±0.004      | 0.977±0.002     | **0.980±0.005** |
|   |    | $U$ as input | 0.940±0.005       | 0.937±0.009      | **0.964±0.003** | 0.961±0.011    |
|   |    | index eigvals | 0.940±0.005      | 0.948±0.001      | 0.982±0.001     | **0.985±0.001** |
|   |    |              | 6 processing steps, 20% node Dropout |               |                  |                   |
| $U$ | GN | spectral MP | 0.938±0.008       | 0.943±0.006      | **0.976±0.005** | nan±nan          |
|   |    | $U$ as input | 0.938±0.008       | 0.946±0.008      | **0.963±0.007** | 0.963±0.005    |
|   |    | index eigvals | 938±0.008        | 0.956±0.001      | **0.979±0.002** | 0.971±0.008    |

Table C.3: Overall best test classification accuracy across 5 seeds given 2.5e5 training steps for different GraphNet architectures. Varying levels numbers of message passing steps.

# D MoleculeNet-MOLHIV

## D.1 Supplementary Methods

MoleculeNet consists of a set of molecular property prediction benchmarks [Wu et al.(2018)Wu, Ramsundar, Feinberg, Gomes, Geniesse, Pappu, Leswing, and Pande]. One of the largest datasets within MoleculeNet is the HIV dataset, in which the challenge is to predict a binary target indicating whether a molecule has an experimentally measured ability to inhibit HIV replication. This benchmark allows us to assess the benefits of spectral augmentation on chemical graph property prediction, a dataset with very different structure to images.

The input data consists of a molecular graph in which nodes are atoms and edges bonds, and node and edge features identify key properties of the atoms and bonds. Both atoms and bonds were encoded as 100 dimensional feature vectors using Open Graph Benchmark's atom and bond encoders, respectively. The dataset consists of 41,127 small, sparse molecular graphs (on average, #nodes=25.5, #edges=27.7, diameter=12.0) and is skewed such that 2.7% of the dataset is labelled positive (see Appendix Figure B.1 for statistics across datasets). This dataset is available on Open Graph Benchmark [Hu et al.(2020)Hu, Fey, Zitnik, Dong, Ren, Liu, Catasta, and Leskovec].

## D.2 Tables and Figures

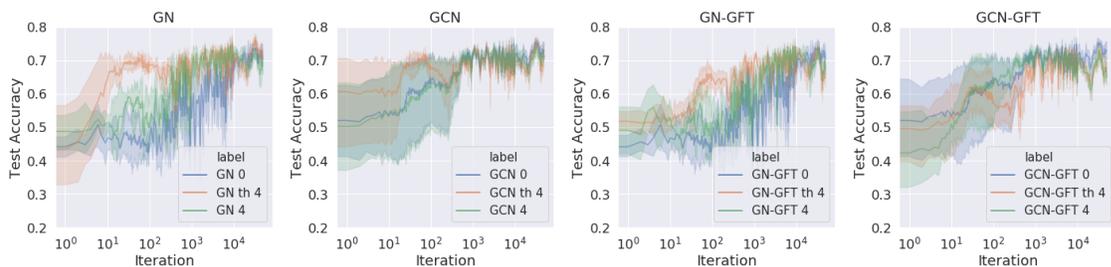

Figure D.6: Model performance over training for GN, GCN, GN-GFT, and GCN-GFT, and their spectral variants, $U$ and $\theta(U)$, on MoleculeNet-HIV.

| | | \multicolumn{4}{c}{MoleculeNet HIV Results} | | | |
|---|---|---|---|---|---|
| | $K$ | 0 | 2 | 4 | 8 |
| $U$ | GCN | 0.755±0.015 | 0.751±0.019 | 0.743±0.021 | **0.761±0.030** |
| | GCN-GFT | **0.755±0.015** | 0.743±0.017 | 0.736±0.017 | 0.741±0.014 |
| | GN | 0.739±0.029 | **0.753±0.028** | 0.747±0.023 | 0.746±0.019 |
| | GN-GFT | 0.739±0.029 | **0.748±0.024** | 0.743±0.024 | 0.742±0.020 |
| $\theta(U)$ | GCN | 0.755±0.015 | 0.755±0.020 | **0.759±0.020** | 0.750±0.032 |
| | GCN-GFT | **0.755±0.015** | 0.754±0.024 | 0.734±0.018 | 0.744±0.014 |
| | GN | 0.739±0.029 | 0.764±0.016 | **0.769±0.015** | 0.748±0.018 |
| | GN-GFT | 0.739±0.029 | **0.758±0.015** | 0.745±0.025 | 0.734±0.014 |

Table D.4: Best performances of each model on MoleculeNet-HIV in terms of ROC-AUC.

| Literature Leaderboard | |
|---|---|
| GatedGCN | 0.7765±0.0050 |
| GIN+virtual node | 0.7707±0.0149 |
| GCN | 0.7606±0.0097 |
| GCN+virtual node | 0.7599±0.0119 |
| GIN | 0.7558±0.0140 |
| Graph-agnostic MLP | 0.6819±0.0071 |

Table D.5: Leaderboard showing the performance of various literature models on MoleculeNet-HIV [Hu et al.(2020)Hu, Fey, Zitnik, Dong, Ren, Liu, Catasta, and Leskovec]

# E  QM9

## E.1  Supplementary Methods

Molecules within the dataset consist of Hydrogen, Carbon, Oxygen, Nitrogen and Fluorine atoms and contain up to 9 heavy (non Hydrogen) atoms. The dataset consists of 134k molecules, of which 10k are randomly selected for validation and test sets. The node features in our input graph consist of atom coordinates, atomic number, formal charge, hybridization, Mulliken particle charge and whether it is aromatic. Edges are defined by the chemical graph, and where applicable edge features of bond type, bond length and bond vector are provided. For targets U0, U, H and G, the reference energy is subtracted. For this task, we train for 5 million steps with a latent size of 32.

## E.2 Tables and Figures

| | | Edge Dropout = 0.0 | | | | |
|---|---|---|---|---|---|---|
| | Unit | GN | GN-$U$ ($K$=2) | GN-$U$ ($K$=4) | GN-$U_{\text{thresh}}$ ($K$=2) | GN-$U_{\text{thresh}}$ ($K$=4) |
| **Cv** | cal/(mol K) | **0.07±0.0** | 0.15±0.0 | 0.15±0.0 | 0.10±0.0 | 0.11±0.0 |
| **G** | meV | **59.35±0.3** | 90.27±27.8 | 61.77±7.8 | 63.81±9.0 | 63.18±6.2 |
| **H** | meV | 69.24±21.2 | 73.89±3.8 | **62.10±0.4** | 79.64±21.8 | 67.03±0.4 |
| **HOMO** | meV | **84.38±0.9** | 109.77±6.4 | 105.53±1.3 | 93.83±1.9 | 101.17±1.0 |
| **LUMO** | meV | **81.19±0.4** | 98.06±1.4 | 103.09±1.9 | 93.16±2.9 | 102.01±6.5 |
| **R2** | Bohr$^2$ | **16.33±0.3** | 27.13±1.3 | 25.78±1.4 | 17.47±0.1 | 17.23±0.1 |
| **U** | meV | 63.28±2.7 | 97.43±23.2 | 68.28±3.5 | **61.79±1.4** | 90.42±6.4 |
| **U0** | meV | **54.81±4.4** | 84.96±3.7 | 64.16±1.7 | 63.19±9.3 | 70.73±17.2 |
| **ZPVE** | meV | **3.79±0.1** | 6.11±0.3 | 5.85±0.4 | 4.12±0.3 | 4.08±0.1 |
| **alpha** | Bohr$^3$ | **0.23±0.0** | 0.39±0.0 | 0.34±0.0 | 0.28±0.0 | 0.28±0.0 |
| **gap** | meV | **126.72±1.8** | 156.22±2.4 | 160.22±0.4 | 145.29±1.9 | 151.85±5.4 |
| **mu** | D | **0.38±0.0** | 0.46±0.0 | 0.47±0.0 | 0.45±0.0 | 0.51±0.0 |
| | | Edge Dropout = 0.5 | | | | |
| | Unit | GN | GN-$U$ ($K$=2) | GN-$U$ ($K$=4) | GN-$U_{\text{thresh}}$ ($K$=2) | GN-$U_{\text{thresh}}$ ($K$=4) |
| **Cv** | cal/(mol K) | **0.12±0.0** | 0.15±0.0 | 0.13±0.0 | 0.14±0.0 | 0.12±0.0 |
| **G** | meV | 127.54±16.7 | 110.92±17.6 | **80.57±4.7** | 119.07±18.4 | 115.22±13.9 |
| **H** | meV | 122.17±6.8 | 118.39±39.6 | **98.61±13.3** | 106.03±1.6 | 124.93±19.0 |
| **HOMO** | meV | **108.92±5.4** | 121.44±6.7 | 114.68±5.0 | 123.66±6.0 | 131.34±1.9 |
| **LUMO** | meV | **107.12±0.3** | 112.95±2.8 | 110.66±0.7 | 115.74±6.1 | 120.44±2.5 |
| **R2** | Bohr$^2$ | **26.70±0.2** | 31.51±0.6 | 28.14±0.1 | 29.18±0.2 | 31.26±0.6 |
| **U** | meV | 118.52±4.4 | 98.12±12.8 | **95.11±9.7** | 113.11±3.2 | 157.62±80.5 |
| **U0** | meV | 124.08±6.8 | 105.79±5.6 | **93.70±3.7** | 121.12±15.8 | 107.42±10.1 |
| **ZPVE** | meV | 5.85±0.2 | 6.31±0.7 | 5.78±0.6 | **5.00±0.4** | 5.25±0.2 |
| **alpha** | Bohr$^3$ | **0.35±0.0** | 0.42±0.0 | 0.41±0.0 | 0.39±0.0 | 0.39±0.0 |
| **gap** | meV | **142.50±1.8** | 179.66±6.4 | 171.94±9.6 | 180.77±1.7 | 185.96±2.3 |
| **mu** | D | **0.46±0.0** | 0.60±0.0 | 0.56±0.0 | 0.59±0.0 | 0.59±0.0 |
| | | Edge Dropout = 1.0 | | | | |
| | Unit | GN | GN-$U$ ($K$=2) | GN-$U$ ($K$=4) | GN-$U_{\text{thresh}}$ ($K$=2) | GN-$U_{\text{thresh}}$ ($K$=4) |
| **Cv** | cal/(mol K) | 0.28±0.0 | 0.21±0.0 | **0.17±0.0** | 0.20±0.0 | 0.21±0.0 |
| **G** | meV | 215.91±5.5 | 168.73±1.2 | **133.99±21.0** | 163.65±22.2 | 145.04±12.4 |
| **H** | meV | 232.32±14.9 | 179.32±1.8 | **125.08±3.2** | 172.41±26.4 | 150.03±2.8 |
| **HOMO** | meV | 140.23±1.1 | 150.29±1.0 | **132.69±1.6** | 146.81±1.4 | 158.18±5.2 |
| **LUMO** | meV | 138.37±0.6 | 134.81±1.2 | **126.81±7.8** | 140.53±4.9 | 141.94±0.0 |
| **R2** | Bohr$^2$ | 56.08±1.0 | 40.89±1.2 | **36.79±2.0** | 45.20±3.1 | 41.28±1.0 |
| **U** | meV | 217.95±5.4 | 166.49±3.6 | **127.26±5.5** | 176.89±30.1 | 152.18±16.9 |
| **U0** | meV | 226.98±8.5 | 170.18±6.6 | **124.98±0.9** | 175.25±1.6 | 150.29±14.8 |
| **ZPVE** | meV | 10.19±0.5 | 8.61±0.0 | **7.51±0.6** | 8.29±0.7 | 7.87±0.3 |
| **alpha** | Bohr$^3$ | 0.58±0.0 | 0.53±0.0 | **0.49±0.0** | 0.52±0.0 | 0.51±0.0 |
| **gap** | meV | 197.88±0.4 | 211.70±2.7 | **189.59±5.6** | 211.23±2.2 | 211.32±9.4 |
| **mu** | D | **0.59±0.0** | 0.66±0.0 | 0.63±0.0 | 0.68±0.0 | 0.65±0.0 |

Table E.6: Mean Absolute Error across $U$ and $\theta(U)$-GN on QM9 targets with varying rates of edge dropout.

|  | | Edge Dropout = 0.0 | | | | |
| --- | --- | --- | --- | --- | --- | --- |
| | Unit | GCN | GCN-$U$ ($K$=2) | GCN-$U$ ($K$=4) | GCN-$U_{\text{thresh}}$ ($K$=2) | GCN-$U_{\text{thresh}}$ ($K$=4) |
| **Cv** | cal/(mol K) | 0.14±0.0 | 0.15±0.0 | 0.15±0.0 | **0.12±0.0** | 0.12±0.0 |
| **G** | meV | 132.11±11.4 | 110.24±1.8 | 111.65±9.0 | **96.79±6.0** | 99.05±7.6 |
| **H** | meV | 142.37±14.1 | 125.69±14.0 | 116.03±3.7 | **100.51±5.3** | 112.59±7.9 |
| **HOMO** | meV | **107.71±2.3** | 117.86±1.8 | 120.90±2.9 | 109.94±1.1 | 113.43±0.5 |
| **LUMO** | meV | **99.75±3.0** | 109.97±2.6 | 110.31±1.7 | 100.01±1.6 | 107.60±1.2 |
| **R2** | Bohr$^2$ | 27.32±1.0 | 30.12±0.1 | 32.37±0.4 | **25.44±0.0** | 27.07±1.0 |
| **U** | meV | 133.49±2.8 | 131.09±12.5 | 143.50±26.4 | **95.45±1.1** | 100.23±3.0 |
| **U0** | meV | 128.15±6.8 | 133.82±23.0 | 112.08±1.6 | 101.20±8.3 | **99.07±13.7** |
| **ZPVE** | meV | 5.60±0.0 | 5.67±0.2 | 5.59±0.2 | 5.42±0.9 | **5.11±0.2** |
| **alpha** | Bohr$^3$ | 0.37±0.0 | 0.39±0.0 | 0.40±0.0 | 0.36±0.0 | **0.35±0.0** |
| **gap** | meV | **143.96±0.6** | 167.64±2.8 | 166.84±3.4 | 153.15±0.7 | 155.86±0.9 |
| **mu** | D | **0.43±0.0** | 0.53±0.0 | 0.53±0.0 | 0.50±0.0 | 0.52±0.0 |
|  | | Edge Dropout = 0.5 | | | | |
| | Unit | GCN | GCN-$U$ ($K$=2) | GCN-$U$ ($K$=4) | GCN-$U_{\text{thresh}}$ ($K$=2) | GCN-$U_{\text{thresh}}$ ($K$=4) |
| **Cv** | cal/(mol K) | 0.32±0.0 | 0.25±0.0 | 0.24±0.0 | 0.20±0.0 | **0.18±0.0** |
| **G** | meV | 330.67±8.2 | 282.96±16.0 | 272.90±2.4 | **166.69±12.9** | 167.59±3.5 |
| **H** | meV | 370.21±5.8 | 280.86±13.6 | 287.68±7.8 | 164.55±2.1 | **161.74±0.7** |
| **HOMO** | meV | 151.60±0.7 | 153.27±0.8 | 154.42±2.8 | **139.58±1.8** | 142.35±3.4 |
| **LUMO** | meV | 144.57±1.8 | 143.03±0.2 | 148.15±0.5 | **131.56±1.2** | 133.95±2.3 |
| **R2** | Bohr$^2$ | 54.69±0.6 | **40.99±1.6** | 41.43±0.3 | 43.90±1.6 | 44.19±2.4 |
| **U** | meV | 365.30±40.5 | 279.27±1.8 | 293.47±15.6 | **167.70±7.9** | 168.76±20.7 |
| **U0** | meV | 340.62±0.7 | 273.01±0.9 | 280.44±13.9 | 164.54±3.3 | **146.67±0.3** |
| **ZPVE** | meV | 14.79±0.3 | 10.43±0.3 | 11.15±0.1 | 7.45±0.2 | **7.35±0.3** |
| **alpha** | Bohr$^3$ | 0.70±0.0 | 0.60±0.0 | 0.61±0.0 | 0.48±0.0 | **0.48±0.0** |
| **gap** | meV | 198.23±2.8 | 202.46±5.3 | 205.60±1.0 | **186.78±0.4** | 188.60±2.8 |
| **mu** | D | **0.54±0.0** | 0.61±0.0 | 0.62±0.0 | 0.59±0.0 | 0.59±0.0 |
|  | | Edge Dropout = 1.0 | | | | |
| | Unit | GCN | GCN-$U$ ($K$=2) | GCN-$U$ ($K$=4) | GCN-$U_{\text{thresh}}$ ($K$=2) | GCN-$U_{\text{thresh}}$ ($K$=4) |
| **Cv** | cal/(mol K) | 0.31±0.0 | 0.26±0.0 | 0.26±0.0 | 0.23±0.0 | **0.22±0.0** |
| **G** | meV | 237.48±7.2 | 191.62±3.5 | 217.30±29.4 | 166.32±14.8 | **147.05±1.7** |
| **H** | meV | 239.66±3.6 | 197.13±1.8 | 202.01±7.5 | 170.63±5.6 | **157.53±1.1** |
| **HOMO** | meV | 148.98±0.6 | 157.66±3.0 | 162.34±0.4 | **140.05±1.0** | 142.38±1.1 |
| **LUMO** | meV | 146.83±3.4 | 156.96±2.5 | 167.63±1.4 | 138.35±1.3 | **138.26±0.6** |
| **R2** | Bohr$^2$ | 58.94±0.5 | **42.20±0.7** | 47.38±0.1 | 46.61±1.4 | 45.23±0.4 |
| **U** | meV | 251.92±5.7 | 197.72±4.3 | 203.85±2.5 | **169.65±2.1** | 172.10±18.7 |
| **U0** | meV | 249.59±7.0 | 208.01±24.1 | 197.44±3.0 | 166.86±10.5 | **154.57±15.7** |
| **ZPVE** | meV | 11.42±0.0 | 8.84±0.2 | 9.35±0.2 | 9.87±0.6 | **7.92±0.4** |
| **alpha** | Bohr$^3$ | 0.60±0.0 | 0.56±0.0 | 0.62±0.0 | 0.51±0.0 | **0.49±0.0** |
| **gap** | meV | 202.90±1.6 | 225.53±1.9 | 235.63±5.7 | 198.51±3.8 | **194.22±4.3** |
| **mu** | D | **0.57±0.0** | 0.65±0.0 | 0.69±0.0 | 0.60±0.0 | 0.60±0.0 |

Table E.7: Mean Absolute Error across $U$ and $\theta(U)$-GN on QM9 targets with varying rates of edge dropout.

|  | Unit | PPGN | SchNet | PhysNet | MEGNet-s | Comorant | DimeNet |
|---|---|---|---|---|---|---|---|
| **Cv** | cal/(mol K) | 0.055 | 0.033 | 0.0529 | 0.05 | 0.13 | 0.0286 |
| **G** | meV | 36.4 | 14 | 9.40 | 12 | - | 8.98 |
| **H** | meV | 36.3 | 14 | 8.42 | 12 | - | 8.11 |
| **HOMO** | meV | 40.3 | 41 | 32.9 | 43 | 36 | 27.8 |
| **LUMO** | meV | 32.7 | 34 | 24.7 | 44 | 36 | 19.7 |
| **R2** | Bohr$^2$ | 0.592 | 0.073 | 0.765 | 0.302 | 0.673 | 0.331 |
| **U** | meV | 36.8 | 14 | 8.15 | 12 | - | 7.89 |
| **U0** | meV | 36.8 | 14 | 8.15 | 12 | - | 8.02 |
| **ZPVE** | meV | 3.12 | 1.7 | 1.39 | 1.43 | 1.98 | 1.29 |
| **alpha** | Bohr$^3$ | 0.131 | 0.235 | 0.0615 | 0.081 | 0.092 | 0.0469 |
| **gap** | meV | 60.0 | 63 | 42.5 | 66 | 60 | 34.8 |
| **mu** | D | 0.047 | 0.033 | 0.0529 | 0.05 | 0.13 | 0.0286 |

Table E.8: Reported Results on QM9 in Literature.

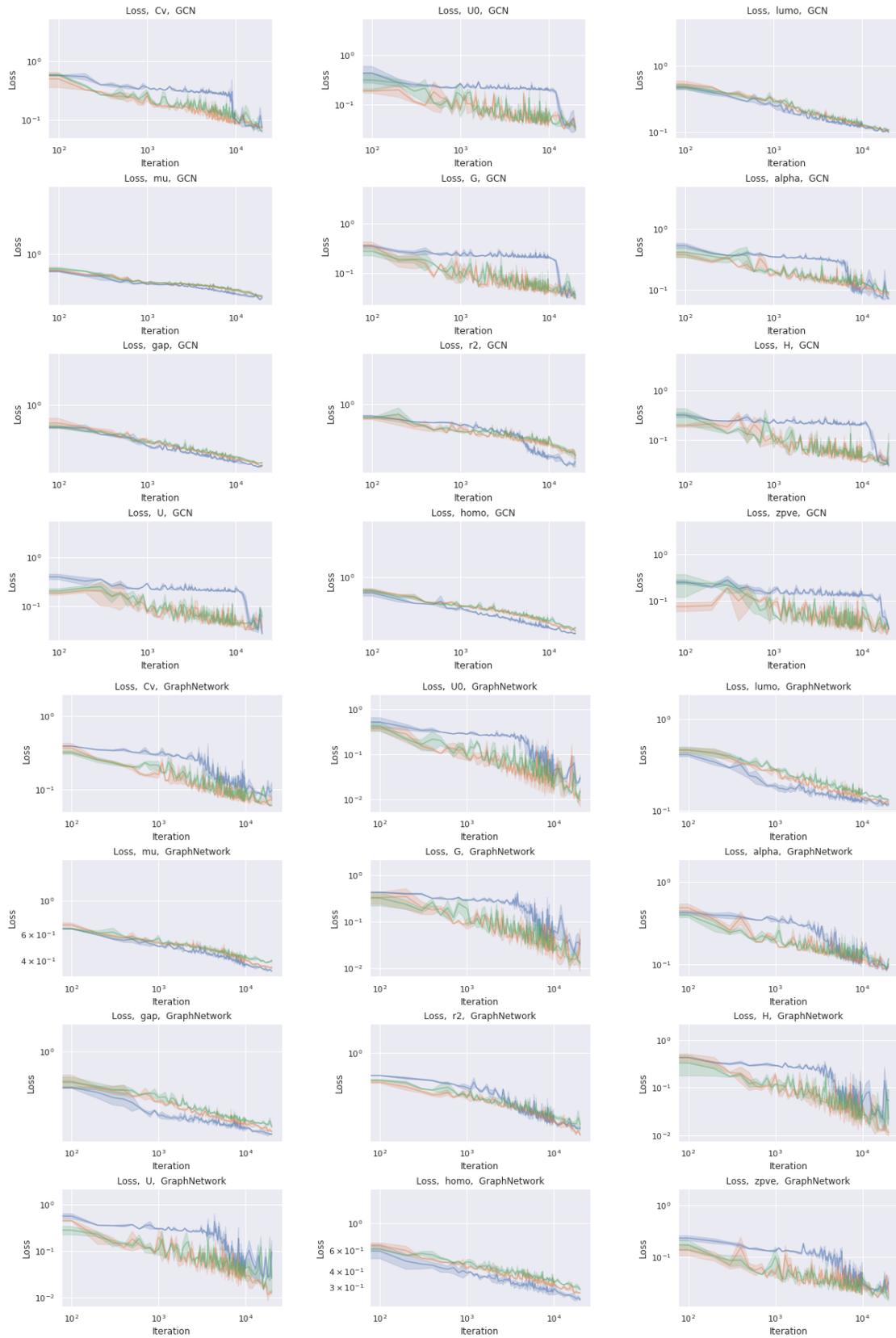

Figure E.7: Early training curves for $\theta(U)$-GCN and $\theta(U)$-GN, with $K=0$ (blue), $K=2$ (orange), and $K=4$ (green).

# F Shortest Path Prediction on Random Graph Datasets

## F.1 Supplementary Methods

This toy benchmark involved randomly sampling graphs from different random graph generators and randomly sampling pairs of nodes as source and target. The pair of nodes was represented as a feature over nodes: the feature was set to 0 if a node was not one of the pair, -1 if it was the source, and 1 if it was the target (although the task in this case did not actually require producing a directed path). The target the model was trained to predict was which nodes would lie on the shortest path joining the two sampled nodes. This was treated as a binary classification problem over nodes, and area under the ROC curve (ROC-AUC) over node classifications was used as the evaluation metric. We chose this metric rather than accuracy partly because some random graphs will have a very small fraction of nodes lie on the shortest path, meaning it was likely that all models would perform highly on accuracy just by setting most node predictions to 0.

Random Delaunay 2D graphs were generated by sampling points uniformly at random from a $1 \times 1$ 2D square and joining them with a Delaunay Triangulation. Edges that connected nodes further than $1.5\times$ the standard deviation of distances for joined nodes were severed to prevent contraction around the boundary. Random Delaunay 2D graphs with obstacles had "obstacles" randomly introduced by sampling a random pair of nodes and removing all nodes along the shortest path joining them.

Barabási-Albert graphs are generated by adding nodes one at a time such that each new node is connected to $m = 1$ existing node with probability proportional to its current degree.

Random Partition graph are a special type of Stochastic Block Model. Nodes were divided into 4 equally sized communities. For two nodes in the same community, the probability of being joined by an edge was 0.7, and for nodes in different communities, it was 0.1.

Our motivation for this analysis was to unpack what properties a dataset might have that would mean that spectral augmentation is the correct inductive bias. Shortest path prediction allowed us to explore the same problem while varying size and statistics of the graphs used.

## F.2 Tables and Figures

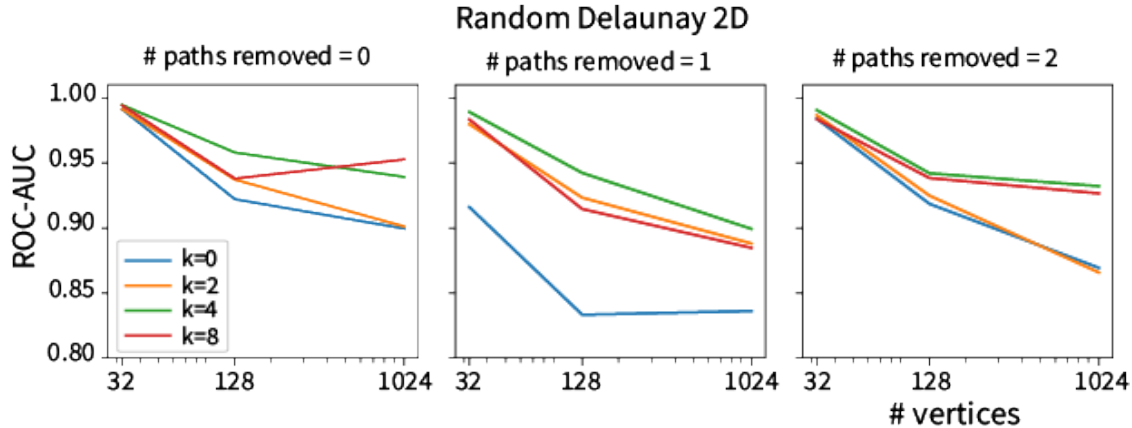

Figure F.8: Effect of number of nodes on performance for different Random Delaunay graphs.

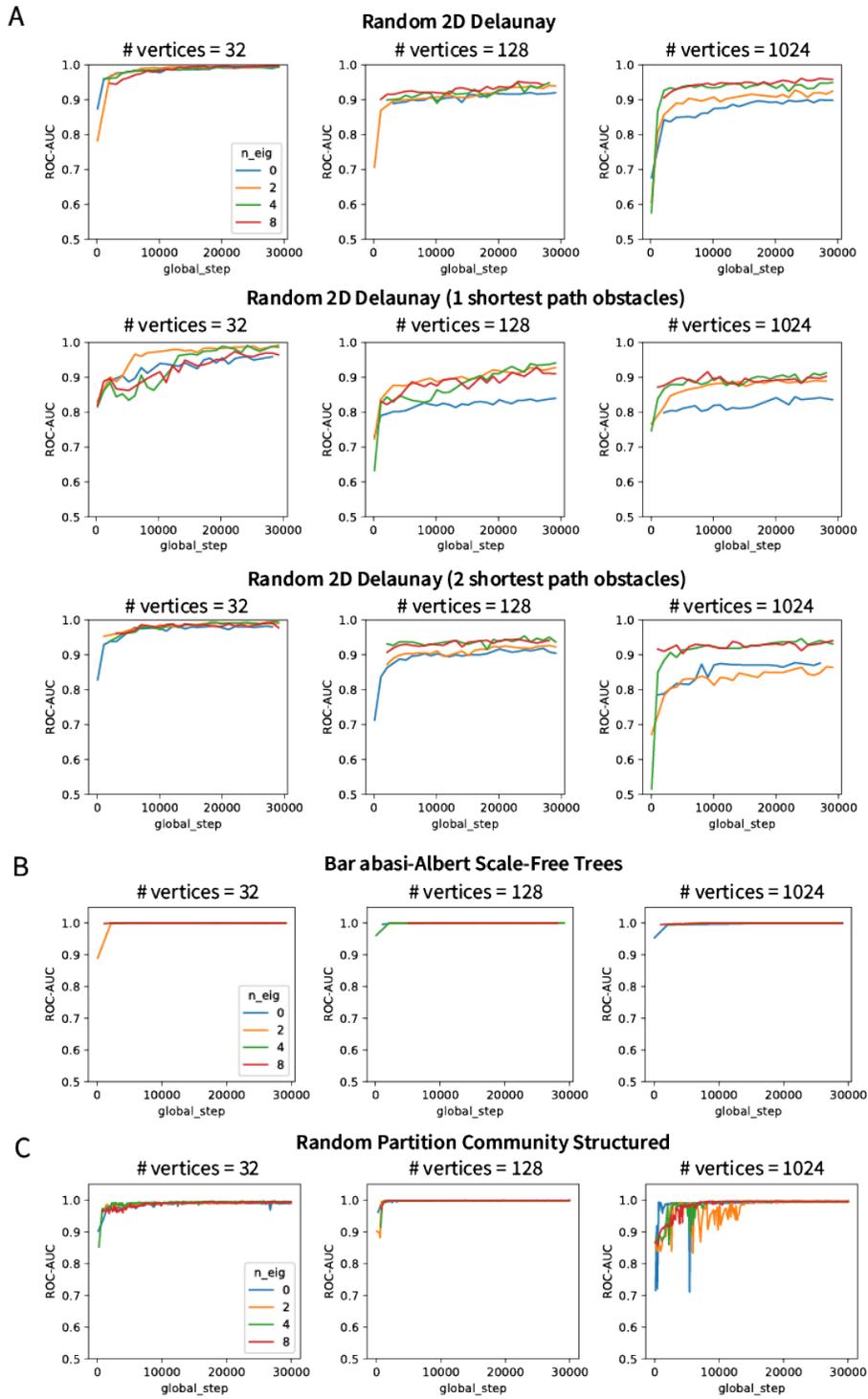

Figure F.9: Model performance over training time for random graph categories and different numbers of nodes. For variants of Random Delaunay 2D graphs, spectral augmentation ($K > 0$) yields increasing performance with number of nodes. However, this trend is not seen for Barabási-Albert and Random Partition Graphs, where spectral augmentation makes no visible difference and actively slows learning, respectively.

18

| Random Delaunay 2D (0 paths removed) | | | | |
| --- | --- | --- | --- | --- |
| $K$ | 0.0 | 2.0 | 4.0 | 8.0 |
| n_node | | | | |
| 32 | 0.991444 | 0.991444 | 0.994652 | 0.994385 |
| 128 | 0.922184 | 0.937225 | 0.95807 | 0.938049 |
| 1024 | 0.899571 | 0.901234 | 0.939284 | 0.95276 |
| Random Delaunay 2D (1 paths removed) | | | | |
| $K$ | 0.0 | 2.0 | 4.0 | 8.0 |
| n_node | | | | |
| 32 | 0.916138 | 0.979862 | 0.989241 | 0.983448 |
| 128 | 0.833367 | 0.923345 | 0.942384 | 0.91464 |
| 1024 | 0.836244 | 0.888102 | 0.899312 | 0.884736 |
| Random Delaunay 2D (2 paths removed) | | | | |
| $K$ | 0.0 | 2.0 | 4.0 | 8.0 |
| n_node | | | | |
| 32 | 0.984074 | 0.986729 | 0.99071 | 0.983743 |
| 128 | 0.918581 | 0.92488 | 0.942105 | 0.938437 |
| 1024 | 0.869434 | 0.865835 | 0.932238 | 0.926654 |
| Barabási-Albert | | | | |
| $K$ | 0.0 | 2.0 | 4.0 | 8.0 |
| n_node | | | | |
| 32 | 1.0 | 1.0 | 1.0 | 1.0 |
| 128 | 1.0 | 0.999766 | 1.0 | 1.0 |
| 1024 | 0.998941 | 0.999231 | 0.999474 | 0.999515 |
| Random Partition | | | | |
| $K$ | 0.0 | 2.0 | 4.0 | 8.0 |
| n_node | | | | |
| 32 | 0.992394 | 0.994524 | 0.993307 | 0.993611 |
| 128 | 0.998903 | 0.998668 | 0.999608 | 0.999138 |
| 1024 | 0.995123 | 0.994864 | 0.996649 | 0.996467 |

Table F.9: Final ROC-AUC for shortest path prediction on different graph datasets.

| | | Random Delaunay 2D | | | |
|---|---|---|---|---|---|
| | $K$ | 0 | 2 | 4 | 8 |
| #vertices | spatial-spectral | | | | |
| 32 | noMP-GN | 0.772727 | 0.923529 | 0.952406 | 0.952674 |
| | GN-noMP | 0.991979 | 0.992781 | **0.997059** | 0.995187 |
| | GN-GN | 0.990909 | **0.994652** | 0.992246 | **0.996257** |
| 128 | noMP-GN | 0.650000 | 0.867960 | 0.900446 | 0.911229 |
| | GN-noMP | 0.886848 | 0.921291 | 0.945398 | 0.952885 |
| | GN-GN | 0.919677 | **0.935405** | **0.948523** | **0.959272** |
| 1024 | noMP-GN | 0.583333 | 0.796574 | 0.904905 | 0.908902 |
| | GN-noMP | 0.907391 | 0.919658 | 0.931176 | 0.945566 |
| | GN-GN | 0.891785 | **0.921260** | **0.949625** | **0.954042** |
| | | Barabási-Albert | | | |
| | $K$ | 0 | 2 | 4 | 8 |
| #vertices | spatial-spectral | | | | |
| 32 | noMP-GN | 0.693548 | 0.932679 | 0.943198 | 0.990383 |
| | GN-noMP | 1.000000 | 1.000000 | 1.000000 | 1.000000 |
| | GN-GN | 1.000000 | 1.000000 | 1.000000 | 1.000000 |
| 128 | noMP-GN | 0.671429 | 0.790879 | 0.908712 | 0.936971 |
| | GN-noMP | 0.999961 | 0.999922 | 1.000000 | 1.000000 |
| | GN-GN | 0.999766 | 1.000000 | 1.000000 | 1.000000 |
| 1024 | noMP-GN | 0.625000 | 0.795287 | 0.862651 | 0.927738 |
| | GN-noMP | 0.999050 | **0.999173** | 0.999231 | **0.999624** |
| | GN-GN | 0.998821 | 0.999101 | **0.999286** | 0.999549 |
| | | Random Partition | | | |
| | $K$ | 0 | 2 | 4 | 8 |
| #vertices | spatial-spectral | | | | |
| 32 | noMP-GN | 0.815789 | 0.888348 | 0.949498 | 0.932765 |
| | GN-noMP | 0.992090 | **0.998175** | **0.997870** | **0.994828** |
| | GN-GN | 0.992699 | 0.991482 | 0.990265 | 0.993003 |
| 128 | noMP-GN | 0.852941 | 0.897940 | 0.977598 | 0.973525 |
| | GN-noMP | 0.999060 | **0.999060** | **0.999295** | 0.998982 |
| | GN-GN | 0.999060 | 0.998512 | 0.998825 | **0.999687** |
| 1024 | noMP-GN | 0.852941 | 0.917304 | 0.955186 | 0.953024 |
| | GN-noMP | 0.994508 | **0.995440** | 0.996419 | **0.996621** |
| | GN-GN | 0.993568 | 0.995267 | **0.996649** | 0.996438 |

Table F.10: Here we show the role of spatial and spectral message passing for shortest path prediction performance as measured by ROC-AUC for different graph datasets. GN indicates GraphNet, NoMP indicates Node features were passed through an MLP (the same MLP for each node) but no message passing was applied. Results suggest that spatial message passing contributes more than spectral, and that for Random Delaunay 2D graphs, spectral message passing (GN-GN) provides a small benefit over spectral processing with no message passing (GN-NoMP).



\end{document}